\documentclass[letterpaper]{article} 
\usepackage{aaai2026}  
\usepackage{times}  
\usepackage{helvet}  
\usepackage{courier}  
\usepackage[hyphens]{url}  
\usepackage{graphicx} 
\usepackage{natbib}  
\usepackage{caption} 
\usepackage{algorithm}
\usepackage{algorithmic}
\usepackage{multirow}
\usepackage{amsfonts}
\usepackage{amssymb}
\usepackage{booktabs}
\usepackage{array}

\usepackage{newfloat}
\usepackage{listings}
\DeclareCaptionStyle{ruled}{labelfont=normalfont,labelsep=colon,strut=off} 
\floatstyle{ruled}
\newfloat{listing}{tb}{lst}{}
\floatname{listing}{Listing}
\usepackage{xspace}
\usepackage{amsmath}
\title{Enabling Delayed-Full Charging Through Transformer-Based Real-Time-to-Departure Modeling for EV Battery Longevity}

\author {
    Yonggeon Lee\textsuperscript{\rm 1},
    Jibin Hwang\textsuperscript{\rm 1},
    Alfred Malengo Kondoro\textsuperscript{\rm 1},
    Juhyun Song\textsuperscript{\rm 2}\footnotemark[1],
    Youngtae Noh\textsuperscript{\rm 1}\thanks{Corresponding authors.}
}

\affiliations {
    \textsuperscript{\rm 1}Hanyang University, Seoul, Republic of Korea\\
    \textsuperscript{\rm 2}KENTECH, Naju-si, Republic of Korea\\
    \textsuperscript{\rm 1}\{yonggeonlee, hjb7165, alfr3do, youngtaenoh\}@hanyang.ac.kr\\
    \textsuperscript{\rm 2}jsong@kentech.ac.kr
}

\newcommand{\ie}{{\emph{i.e.},}\xspace}
\newcommand{\eg}{{\emph{e.g.},}\xspace}
\newcommand{\ours}{\textit{DFC}\xspace}

\begin{document}

\maketitle

\begin{abstract}
Electric vehicles (EVs) are key to sustainable mobility, yet their lithium-ion batteries (LIBs) degrade more rapidly under prolonged high states of charge (SOC). This can be mitigated by delaying full charging \ours until just before departure, which requires accurate prediction of user departure times. In this work, we propose Transformer-based real-time-to-event (TTE) model for accurate EV departure prediction. Our approach represents each day as a TTE sequence by discretizing time into grid-based tokens. Unlike previous methods primarily dependent on temporal dependency from historical patterns, our method leverages streaming contextual information to predict departures. Evaluation on a real-world study involving 93 users and passive smartphone data demonstrates that our method effectively captures irregular departure patterns within individual routines, outperforming baseline models. These results highlight the potential for practical deployment of the \ours algorithm and its contribution to sustainable transportation systems. The code is available
at https://github.com/LYGLeo/3TD-AISI-26.
\end{abstract}

\section{Introduction}
\label{sec:intro}

\begin{figure}[t]
\centering
\includegraphics[width=0.95\columnwidth]{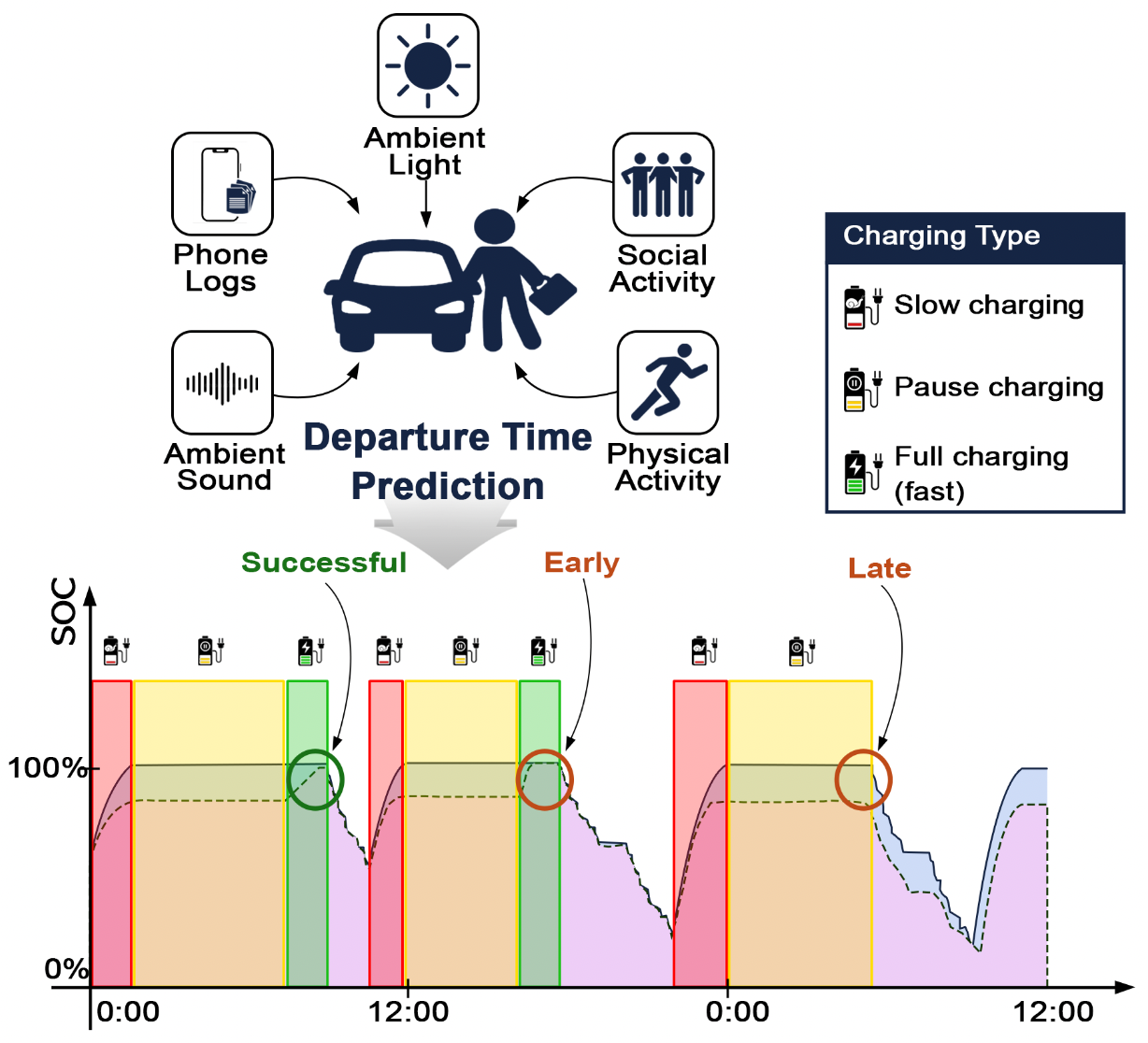} 
\caption{Virtual operation of \ours. Early predictions cause prolonged high SOC, reducing battery health benefits. Late predictions risk insufficient charge and range anxiety. Accurate prediction of departure at least 30 minutes in advance ensures full charging for effective \ours operation, extending battery life while maintaining user confidence.}
\label{fig1}
\end{figure}

As global efforts toward net-zero emissions intensify, electric vehicles (EVs) have emerged as a key component of sustainable transportation. However, large-scale adoption is hindered by lithium-ion battery (LIB) degradation, which reduces driving range, raises replacement costs, and leads to environmental waste. A primary cause of degradation is extended exposure to high state of charge (SOC), often resulting from early, unnecessary full charging before use. To address this, the Delayed-Full Charging (\ours) strategy was proposed in~\cite{lee2024extending}. \ours postpones charging until shortly before the expected EV departure time (\eg 30 minutes), then performs fast charging to reach full SOC, thereby minimizing the dwell time at 100\% SOC ($t_{100\%}$). This approach promotes sustainability by extending battery life and aligns with emerging global regulations, such as the European Union’s 2027 mandate on battery health.

Indeed, the practical efficacy of the \ours algorithm relies on accurate prediction of EV departure time. Inaccurate predictions introduce two critical risks: early charging reduces \ours’s benefits by prolonging high SOC exposure, whereas late charging can leave insufficient SOC at departure, causing user discomfort (Figure~\ref{fig1}). Consequently, departure time emerges as a key control parameter for charging scheduling within the \ours framework. Existing methods primarily rely on historical statistics of usage patterns, offering limited adaptability to dynamic user contexts~\cite{frendo2021smart, lindroth2024online}. These approaches often overlook real-time causal signals of departure, leading to poor performance and interpretability. For robust deployment of \ours, predictive models must provide temporal precision while adapting to contextual signals that reflect pre-departure behavior.

Pre-departure behavior is shaped by diverse contextual cues, including user behavioral signals such as physical activity, smartphone engagement, and social interactions, as well as environmental factors like ambient light and sound. Smartphone passive sensing provides a scalable and unobtrusive way to capture these signals continuously, enabling a digital phenotyping approach aligned with the quantified-self paradigm~\cite{torous2016new}. Prior work has shown that machine learning classifiers trained on sliding-window contextual features can improve first-daily departure prediction from home. Combined with DBSCAN, these models outperform historical-pattern baselines~\cite{lee2024extending}. Yet, framing departure prediction as classification leads to severe label imbalance, which becomes more pronounced at finer time resolutions for real-time inference.

Building on recent advances in deep survival analysis across domains such as healthcare, churn prediction, and mobility, we reformulate departure time prediction as a time-to-event (TTE) problem, applying survival analysis principles to estimate the probability distribution of departure. This reframing offers two key advantages: (1) it mitigates label imbalance by representing each day as a single observation instead of multiple binary windows, and (2) it supports real-time inference where the event time is unknown until departure, enabling probabilistic predictions. Unlike classification, TTE models predict survival functions, providing richer temporal modeling of departure likelihood. To this end, we propose a time-to-departure (TTD) modeling approach based on a Transformer architecture, leveraging its ability to capture temporal dependencies and achieve efficient parallel processing through multi-head attention. The model updates survival probabilities over discretized time grids using real-time contextual signals from passive sensing. This enables timely departure prediction and providing a robust foundation for successful deployment of \ours in real-world environments. Our contributions are threefold:

\begin{itemize}
    \item We introduce the departure time prediction problem and formulate it as a TTE prediction task. This enables \ours\ to effectively reduce $t_{100\%}$ and extend battery lifetime.
    \item We design a Transformer-based TTD framework that enables token-wise streaming inference, incorporating an ordinal Gaussian-smoothed loss and regularization strategies for robust learning under dynamic contexts.
    \item We validate our method on a real-world dataset of 93 users, achieving significant performance gains over the baselines, and release the dataset for future research.
\end{itemize}

This work takes a first step toward establishing benchmark models and datasets for smart EV charging strategies that depend on accurate departure prediction, such as \ours. Our approach offers strong potential for social impact by reducing battery degradation and supporting sustainable EV usage in response to climate challenges.

\section{Related Work}
\subsection{Departure Time Prediction}
Departure time prediction problem is often treated as part of broader mobility modeling objectives, such as EV behavior prediction and trajectory forecasting. 

In the EV domain, prior research has primarily focused on predicting charging-related behaviors, especially departure and arrival times. Most approaches leverage historical data and machine learning to forecast departure times from temporal usage trends.~\cite{frendo2021smart} propose a regression-based framework for workplace charging, integrating predictions into a heuristic scheduling strategy. To capture sequential dependencies, ~\cite{khwaja2021lstm} and ~\cite{boulakhbar2022ev} employ LSTM networks, targeting residential charging and e-taxi operations, respectively. Recently,~\cite{lindroth2024online} introduce online learning models for real-time departure and distance prediction, adapting to driving behaviors. However, existing methods largely rely on historical patterns without incorporating dynamic contextual signals influencing departure decisions.

Deep learning methods for mobility prediction largely fall into two branches: Recurrent Sequence Models and Neural Temporal Point Process (TPP) Models. \textbf{Recurrent Sequence Models.} ST-RNN~\cite{liu2016predicting} extends RNNs with time- and distance-specific transition matrices to model local spatiotemporal contexts for next-location prediction. DeepMove~\cite{feng2018deepmove} employs historical attention and multi-modal embeddings to capture long-term, periodic mobility patterns. \textbf{Neural TPP Models.} RMTPP~\cite{du2016recurrent} models event timing in continuous time via RNN-based intensity functions, and RSTPP~\cite{yang2018recurrent} extends this with spatial context for check-in time prediction. TrajTPP~\cite{zeng2025inquiry} introduces dual attention and spatio-temporal GRUs for joint location and travel-time prediction. Recent advances integrate graph structures and attention mechanism, as in STGNPP~\cite{jin2023stgnpp} for congestion event prediction and TANTPP~\cite{zhang2024multi} for train delay forecasting.

While these approaches achieve robust performance in trajectory and event-sequence prediction, individual departure time prediction remains underexplored as a standalone task. Prior models leverage historical trajectories and inter-event relations, whereas our work focus on real-time contextual features for dynamic departure detection. As baseline, we employ state-of-the-art sequential architectures (\eg iTransformer~\cite{liu2023itransformer}) to learn temporal or contextual dependencies from behavioral sequences.

\subsection{Deep Survival Analysis}
Early deep learning approaches for survival analysis extended the Cox proportional hazards (PH) framework. DeepSurv~\cite{katzman2018deepsurv} replaced the linear predictor with a neural network for non-linear risk estimation. Kvamme et al.~\cite{kvamme2019time} improved scalability and introduced non-proportional hazard extensions. DeepEH~\cite{zhong2021deepeh} unified CoxPH and AFT using a deep extended hazard formulation to account for environment-specific effects. DiffSurv~\cite{vauvelle2023diffsurv} reframed survival analysis as a differentiable ranking problem for concordance optimization. 

In contrast, non-parametric models discard the proportional hazards assumption and directly learn the survival distribution as a flexible function. DeepHit~\cite{lee2018deephit} modeled discrete-time survival as a multi-task classification for competing risks. SurvPP~\cite{kim2023survpp} used a permanental point process to estimate continuous-time hazards with time-varying covariates. Beyond architectural innovations, recent research has focused on improving robustness and fairness under distribution shifts. Hu and Chen~\cite{hu2024fairness} proposed a distributionally robust framework to mitigate subgroup disparities. Stable Cox~\cite{fan2024stablecox} reweights samples to identify stable predictors. In the mobility domain, RCR~\cite{yang2018rcr} applies recurrent-censored regression with RNNs to predict user check-in times by capturing spatiotemporal dependencies from past trajectories, while DILSA~\cite{vahedian2019predicting} proposes a two-stage deep survival framework to predict abnormal taxi demand surges within a five-hour horizon.

Despite recent advances in deep survival analysis~\cite{lee2019dynamic, wang2022survtrace, mesinovic2024dysurv, zhang2025transformerlsr}, existing methods either violate covariate assumptions, learning approach, or introduce auxiliary objectives that increase complexity. In this work, we focus on data-driven real-time departure detection using exogenous contextual features. Prior survival models are not directly comparable, as their theoretical foundations and architectures do not align with our setting (see supplementary material for details). Therefore, we build on the Transformer-based deep survival framework~\cite{hu2021transformer}, which models survival probabilities over discretized time grids using ordinal regression. We extend this design with contextual feature embeddings and token-level survival updates, enabling streaming inference for timely, accurate departure detection.

\section{Methodology}

\subsection{Problem Definition}
We formulate real-time EV departure detection as a TTE prediction problem, specifically as a TTD task. Let a day be discretized into $T_{\max}$ intervals (\ie 5 minutes), indexed by $t \in \{0,1,\dots,T_{\max}\}$. For a given user-day sequence, the contextual feature at time $t$ is denoted $\mathbf{x}_t \in \mathbb{R}^d$, capturing smartphone-based sensing signals and temporal indicators such as time-of-day and day-of-week. The partial observation up to time $t$ is $\mathcal{X}_{0:t} = \{\mathbf{x}_0,\dots,\mathbf{x}_t\}$. We focus on the first significant departure of the day, which typically corresponds to the EV unplug event after long-duration charging. Let $t^{*} \in \{0,\dots,T_{\max}\}$ denote this event time, defined as the interval 30 minutes prior to the actual departure to ensure sufficient time for fast charging to reach 100\% SOC.

We assume a guaranteed daily departure in our target scenario, and thus all training samples are treated as uncensored. During inference, however, the model observes only partial sequences for which the event has not yet occurred, so the true event time remains unknown and each sequence is right-censored at the current time $t$.

\subsection{Survival Analysis for Time-to-Departure}
Survival analysis provides a statistical framework for modeling TTE outcomes under uncertainty and right-censoring. Let $T$ denote the random variable representing the departure time, and let $\mathbf{X}$ denote the observed covariates. The event distribution is characterized by the survival function
\[
S(t \mid \mathbf{X}) = P(T > t \mid \mathbf{X}),
\]
which gives the probability that departure has not occurred by time $t$. In discrete time, the hazard function is
\[
\lambda(t \mid \mathbf{X}) = P(T = t \mid T \ge t, \mathbf{X}),
\]
representing the probability of departure at interval $t$ given survival up to $t-1$. Defining the hazard complement $q(t \mid \mathbf{X}) = 1 - \lambda(t \mid \mathbf{X})$, the survival function becomes
\[
S(t \mid \mathbf{X}) = \prod_{\tau=0}^{t} q(\tau \mid \mathbf{X}).
\]
This discrete-time formulation characterizes the relationship between hazard and survival functions. In our model, we directly estimate the survival probability at each interval, denoted $\hat{S}(t \mid \mathbf{X})$, rather than predicting hazards explicitly.

\subsection{Transformer-Based Time-to-Departure Modeling}
\label{sec:3.3}
Our framework predicts $\hat{S}(t \mid \mathbf{X})$ for all intervals in parallel using a Transformer-based architecture, enabling efficient survival estimation with real-time updates. Through self-attention, the model captures both regular daily routine and dynamically changing contextual patterns, supporting timely and adaptive departure predictions~\cite{hu2021transformer}.

\begin{figure}[t]
\centering
\includegraphics[width=1\columnwidth]{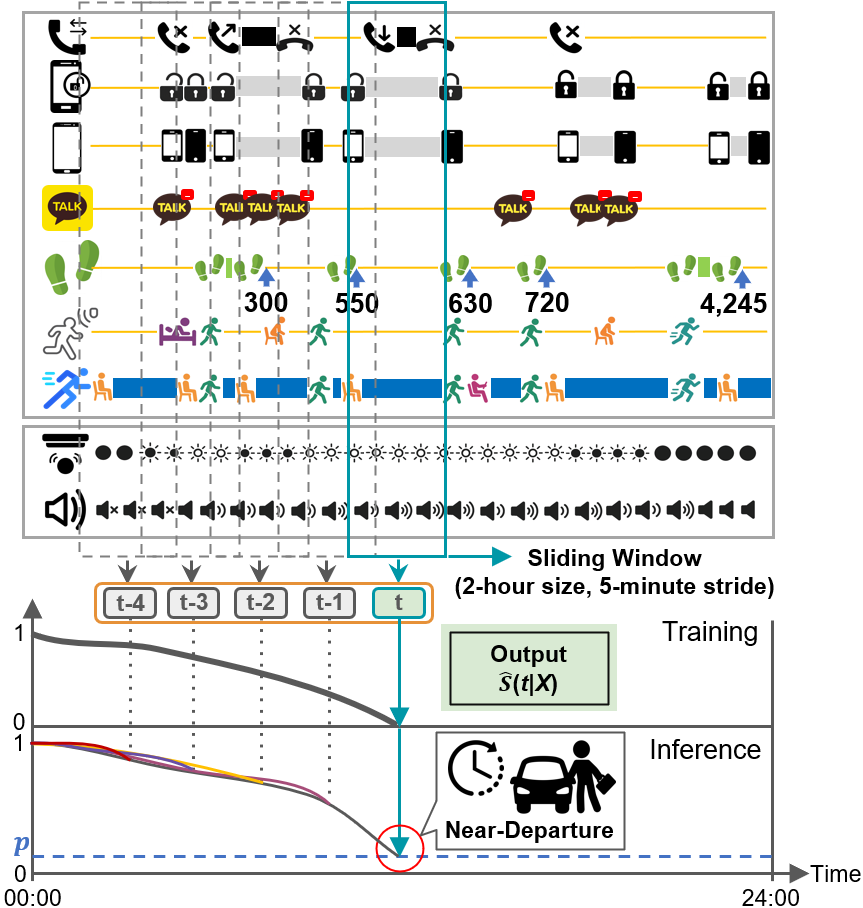} 
\caption{Real-time survival modeling with streaming contextual inputs. During training, the model estimates survival probabilities up to the observed departure event for uncensored sequences. At inference, survival estimates are updated incrementally as new tokens arrive, triggering a departure once the probability falls below a threshold $p$.}
\label{fig2}
\end{figure}

\vspace{4pt}
\noindent\textbf{Objective.}
Given contextual observations up to time $t$, $\mathcal{X}_{0:t}$, the goal is to estimate the sequence of survival probabilities 
$\{\hat{S}(\tau \mid \mathbf{X})\}_{\tau=0}^{T_{\max}}$, which characterizes the likelihood that departure has not occurred by each interval. 
Formally, the model produces
\[
\hat{S}_{0:T_{\max}} = f_\theta(\mathcal{X}_{0:t}),
\]
where $f_\theta$ is the Transformer-based architecture parameterized by $\theta$. 
During training, supervision is applied only up to the observed departure time $t^{*}$, enforcing survival for $t < t^{*}$ and failure at $t = t^{*}$. 
During inference, the model processes partial sequences in real time, updating survival estimates and triggering departure detection once the predicted survival probability falls below a decision threshold~$p$. Figure~\ref{fig2} illustrates this real-time update process.

\vspace{4pt}
\noindent\textbf{Input Representation.} Each interval $t$ is represented by a token integrating three components: 
(1) contextual features from passive smartphone sensing (\eg activity transitions, ambient environment), 
(2) absolute time features encoding the interval’s position in the day, and 
(3) day-of-week features capturing weekday–weekend (and holiday) variation. 
Contextual and day-of-week features are concatenated and projected into a shared embedding space with layer normalization and dropout. 
Absolute time features are projected through a non-linear feed-forward network $g(\cdot)$ and scaled by a learnable parameter $\gamma \in \mathbb{R}$. 
The time embedding is
\[
\mathbf{z}^{\text{time}}_t = \gamma \cdot g(\text{abs\_time}_t).
\]
Let $\alpha \in (0,1)$ be a learnable parameter controlling the balance between contextual and temporal signals. The fused token embedding is then
\[
\mathbf{h}_t = \alpha \cdot \mathbf{z}^{\text{context}}_t + (1-\alpha)\cdot \mathbf{z}^{\text{time}}_t.
\]
All token embeddings share a latent space of size $d_{\text{model}}$ within the Transformer encoder.

\vspace{4pt}
\noindent\textbf{Sequence Modeling.} To encode temporal order, sinusoidal positional encodings are added to the fused embeddings. The resulting sequence $\{\mathbf{h}_0,\dots,\mathbf{h}_{T_{\max}}\}$ is processed by a multi-layer Transformer encoder composed of self-attention and feed-forward sublayers. Unlike recurrent survival models, this design processes all time intervals in parallel, improving efficiency and enabling flexible partial-day inference.

\vspace{4pt}
\noindent\textbf{Output Layer.}
For each interval $t$, the Transformer produces an output representation $\mathbf{u}_t \in \mathbb{R}^{d_{\text{model}}}$, which is passed through a feed-forward network with a sigmoid activation:
\[
\hat{S}(t \mid \mathbf{X}) = \sigma(W_o \mathbf{u}_t + b_o),
\]
where $\hat{S}(t \mid \mathbf{X}) \in (0,1)$ denotes the predicted survival probability $P(T > t \mid \mathbf{X})$. These per-interval estimates form a discrete survival curve that updates in real time as contextual observations evolve throughout the day.

\noindent\textbf{Regularization Strategies.}
To improve robustness against over-reliance on temporal patterns, we introduce three mechanisms:
(1) dropout-time, which randomly masks absolute time features during training;
(2) time-scale, a learnable scalar that moderates the strength of temporal position embeddings; and
(3) alpha-fusion, which balances contextual and temporal representations through a learnable weight.
These strategies regularize temporal bias and promote generalization across diverse behavioral patterns.

\subsection{Loss Function}
We optimize a discrete-time ordinal regression objective inspired by~\cite{hu2021transformer}. 
Let the training set contain $N$ user-day sequences indexed by $i \in \{1, \dots, N\}$. 
For each sequence $i$, let $\mathbf{X}_i$ denote the full contextual feature sequence and 
$t^*_i \in \{0, \dots, T_{\max}\}$ denote the observed (uncensored) departure time. 
The model outputs per-interval survival probabilities $\hat{S}(t \mid \mathbf{X}_i)$ for 
$t \in [0, T_{\max}]$. Unlike~\cite{hu2021transformer}, which parameterizes the hazard complement, we directly model $S(t)$ to avoid unnecessary intermediate variables, obtain cleaner uncertainty estimates, and simplify downstream inference for real-time departure detection. 
This survival estimate forms the basis for the ordinal regression loss used to supervise the model up to the observed event time $t_i^*$.

\vspace{4pt}
\noindent\textbf{Ordinal Regression Loss.}
Since all training sequences contain an observed departure event, we optimize the negative log-likelihood for uncensored samples. 
The ordinal regression loss encourages the model to assign high survival probability before the event and a sharp probability drop at the event time:
\[
\mathcal{L}_{\text{ord}, i}
= - \sum_{t=0}^{t^{*}_i-1} \log \hat{S}(t \mid \mathbf{X}_i)
  \;-\; \log \bigl( 1 - \hat{S}(t^{*}_i \mid \mathbf{X}_i) \bigr).
\]
The first term encourages the user to remain in the ``not departed'' state for all intervals before $t^{*}_i$, while the second enforces a transition to the departure state at $t^{*}_i$. 
Predictions after the event are excluded, as they are not meaningful in TTD modeling. 
Since non-departure days represent true non-events rather than right-censored sequences, training is performed solely on samples where a departure is observed.

\vspace{4pt}
\noindent\textbf{Soft Supervision with Gaussian Smoothing.}
Short adjacent intervals (\ie 5 minutes) may share similar context, making strict supervision sensitive to minor timing shifts near the event. 
To mitigate this, we apply a normalized Gaussian weighting kernel centered at $t^*_i$, restricted to pre-event intervals, to smooth the loss around the departure time. 
This Gaussian-Smoothed Supervision (GSS) enhances robustness to label uncertainty and contextual noise immediately preceding the event:

\[
w_i(t)
= \frac{\exp\bigl[-(t - t^*_i)^2 / (2\sigma^2)\bigr]}
       {\sum_{\tau \le t^*_i} \exp\bigl[-(\tau - t^*_i)^2 / (2\sigma^2)\bigr]},
\qquad t \le t^*_i.
\]
The smoothed ordinal regression loss becomes
\[
\begin{aligned}
\mathcal{L}_{\text{smooth}, i}
= - \sum_{t=0}^{t^{*}_i} w_i(t) \Big[
& \mathbf{1}_{t < t^*_i} \log \hat{S}(t \mid \mathbf{X}_i) \\
& {}+ \mathbf{1}_{t = t^*_i} \log \bigl(1 - \hat{S}(t \mid \mathbf{X}_i)\bigr)
\Big].
\end{aligned}
\]

\vspace{4pt}
\noindent\textbf{Event and Weekend Weights.}
We incorporate two weighting factors: (1) an event weight $\omega_e$, which amplifies the failure term at $t^{*}_i$ and scales the corresponding gradients; and 
(2) a weekend weight $\omega_w$, which adjusts sequence-level contributions to account for irregular weekend patterns and stabilize the overall loss.

\vspace{4pt}
\noindent\textbf{Total Loss.}
The total loss aggregates the per-sequence losses with temporal and day-type weighting:
\[
\mathcal{L}_{\text{total}}
= \sum_{i=1}^{N} \omega_w^{(i)}
\Bigg(
\sum_{t=0}^{T_{\max}} w_i(t)\,\ell_{i,t}
\Bigg),
\]
\[
\text{where } \ell_{i,t} =
\begin{cases}
- \log \hat{S}(t \mid \mathbf{X}_i), & t < t^{*}_i, \\[4pt]
- \omega_e \log \bigl(1 - \hat{S}(t^{*}_i \mid \mathbf{X}_i)\bigr), & t = t^{*}_i, \\[4pt]
0, & t > t^{*}_i,
\end{cases}
\]
\[
\text{and } \omega_w^{(i)} =
\begin{cases}
\omega_w, & \text{if weekend/holiday}, \\[4pt]
1, & \text{otherwise (weekday)}.
\end{cases}
\]
This composite loss enforces survival consistency before $t^{*}_i$, sharp supervision at the event time, and robustness to day-type heterogeneity.

\begin{figure}[t]
\centering
\includegraphics[width=1\columnwidth]{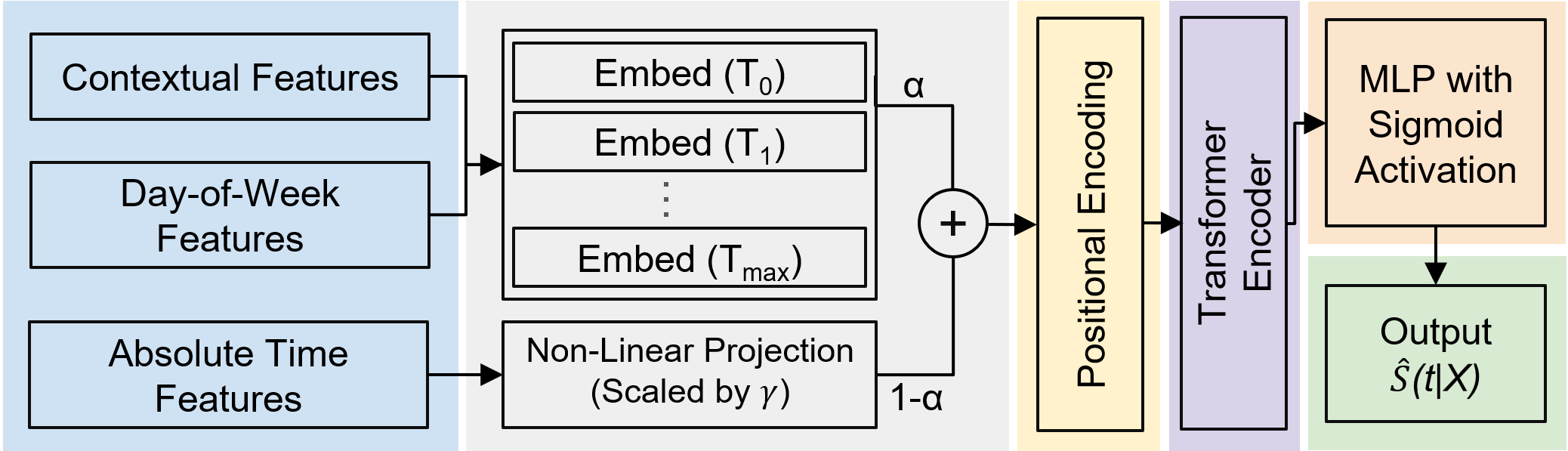}
\caption{A diagram of the TTD model architecture. Contextual features and day-of-week embeddings are fused with absolute time features through the alpha-fusion mechanism, with absolute time scaled by a learnable parameter to control its contribution relative to contextual signal. The fused embeddings, combined with positional encoding, are processed by a multi-layer Transformer encoder to capture long-range temporal dependencies. The output layer predicts per-interval survival probabilities $\hat{S}(t \mid \mathbf{X})$ across all intervals in parallel during training and supports real-time inference by updating predictions as new tokens arrive.}
\label{fig3}
\end{figure}

\section{Experiments}
\label{experiments}

\subsection{Dataset}
\label{sec:dataset}

\textbf{Participant Recruitment.} 
We conducted an IRB-approved field study between May 2021 and July 2022, recruiting 506 Android users (ages 18–69) from the Seoul metropolitan area through social media and community postings. Given the low prevalence of EV ownership, we targeted general smartphone users under the assumption that daily departure routines are behaviorally consistent across populations.

\vspace{4pt}
\noindent\textbf{Passive Sensing and Departure Labels.}
We developed \textit{EV Analyzer} (\textit{EVA}), an Android application for background passive sensing that collects nine behavioral and environmental streams (\ie activity transitions, step counts, significant motion, screen state, unlock state, app usage, call logs, ambient light, and ambient sound). Data were captured through a combination of periodic sampling (\eg light every 15 minutes) and event-driven triggers. Ground-truth departure times were derived from activity transition logs by identifying the first occurrence of the ``in-vehicle'' activity each day.

\vspace{4pt}
\noindent\textbf{Data Preprocessing.}
To ensure data reliability, we applied three steps: (1) participant filtering based on data quality, (2) exclusion of temporally inconsistent records, and (3) removal of values outside realistic ranges. The final dataset comprises 93 participants with 42 days each, totaling 3{,}906 daily sequences.

\vspace{4pt}
\noindent\textbf{Feature Extraction.}
We extracted features using statistical summaries for continuous signals and count- or duration-based metrics for event-driven logs. Additionally, we incorporated entropy measures and elapsed-time features that capture the time since the first occurrence of each key event. To determine the optimal sliding-window size, we conducted Spearman rank correlation analysis with statistical significance test. Features were computed within this window and updated every 5 minutes, resulting in 98 features for the Transformer-based TTD model. Additional correlation results are provided in the supplementary material.

\vspace{4pt}
\noindent\textbf{Justification and Availability.}
This dataset is unique in providing fine-grained, real-world behavioral traces aligned with accurately labeled departure events, a combination not available in existing benchmarks. Since no comparable public dataset exists, custom data collection was required. To support reproducibility and future research, the anonymized dataset and full processing pipeline will be released upon publication (see supplementary material).

\subsection{Experimental Setups}
\label{sec:4.2}

\textbf{Training Details.}  
We train a generalized global model by splitting participants in a 4:1 ratio for training and testing. The training set is further split into training and validation subsets (4:1), yielding 60 users for training, 15 for validation, and 18 for testing. Model and training hyperparameters (\eg number of attention heads, learning rate) are tuned via grid search with 5-fold cross-validation. 

The TTD model uses a 3-layer Transformer encoder with $d_{\text{model}}=32$ and a single attention head, followed by a task-specific output layer. Training employs the Adam optimizer with early stopping and a maximum of 100 epochs. All experiments are implemented in PyTorch and run on an NVIDIA RTX 4090 GPU. For robustness, each experiment is repeated with three random seeds (\ie 42, 43, 44), and we report the mean and standard deviation across runs. Additional implementation details, including hyperparameter ranges, are provided in the supplementary material.

\vspace{4pt}
\noindent\textbf{Evaluation Metric.}
Performance is measured using mean absolute error (MAE) in hours, which provides an intuitive measure of temporal deviation. The predicted departure time is taken as the earliest time index at which the estimated survival probability falls below a decision threshold $p$. We vary this $p$ from 0.05 to 0.20 in increments of 0.05 to assess its impact on accuracy, with 0.10 selected as the optimal value through hyperparameter tuning.

\begin{figure}[t]
\centering
\includegraphics[width=1\columnwidth]{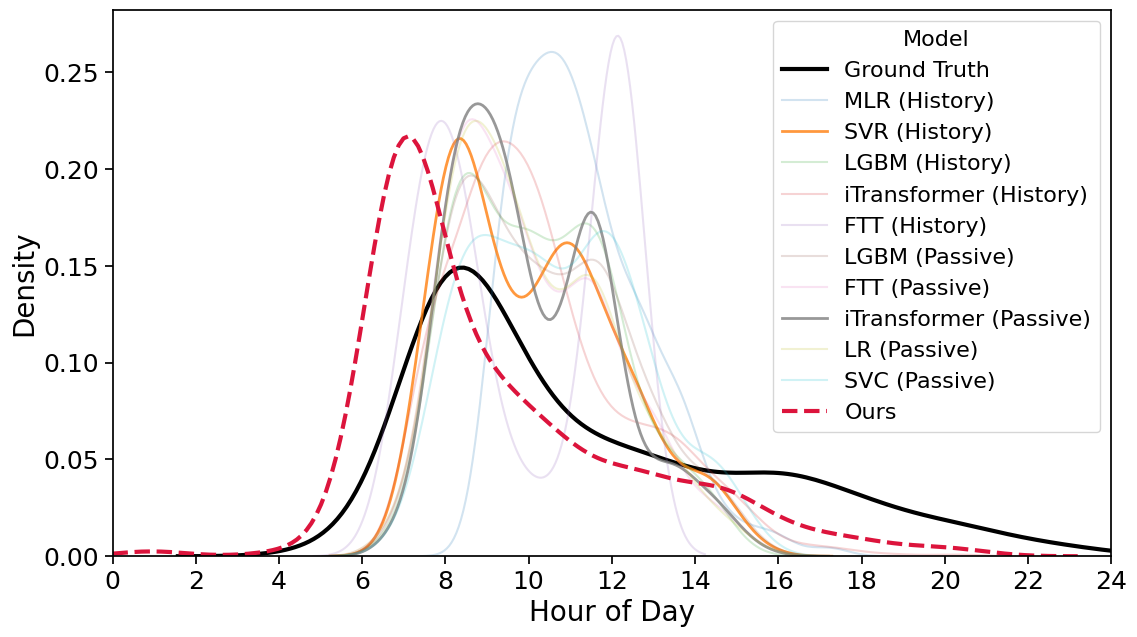} 
\caption{Kernel Density Estimation (KDE) plots of departure times across models. The proposed model shows the closest alignment with the ground-truth distribution compared to the best regressor and classifier baselines.}
\label{fig4}
\end{figure}

\subsection{Baseline Comparison}
\label{sec:baseline}
We benchmark our TTD model against two baselines: (1) a regressor trained on historical statistics, and (2) a context-aware classifier combined with DBSCAN~\cite{lee2024extending}.

\vspace{4pt}
\noindent\textbf{Historical Statistics-Based Regressors.} These baselines are trained with statistics of observed departure times from the previous seven days: mean, standard deviation, minimum, maximum, kurtosis, and skewness, which are computed separately for all days, weekdays, and weekends, with an additional binary day-of-week indicator (19 features in total). We evaluate five models on this feature set: multiple linear regression (MLR), support vector regression (SVR), LightGBM Regressor, FT-Transformer (FTT~\cite{gorishniy2021revisiting}) and iTransformer in a regression setting.

\vspace{4pt}
\noindent\textbf{Context-Aware Classifiers.} Following the strategy in~\cite{lee2024extending}, these baselines use the same contextual features as our method and employ the same set of model architectures in a classification setting. We further incorporate quantile regression to estimate a prediction boundary and apply DBSCAN to detect sharp probability rises, triggering departure detection only within the predicted range.

\vspace{4pt}
\noindent\textbf{Performance Evaluation.} Table~\ref{tab:baseline_comparison} reports MAE (hours) across all days, weekdays, and weekends. Our model achieves the lowest error (2.20 hours), outperforming the best historical baseline (SVR, 2.57 hours) and the context-aware classifier (iTransformer, 2.59 hours). This corresponds to relative improvements of 14.4\%/15.1\% overall, 10.3\%/11.7\% on weekdays, and 21.3\%/21.9\% on weekends, demonstrating the effectiveness of modeling departure as a survival process rather than relying on point prediction or classification-based approaches. Notably, weekday and weekend performance is comparable across all models, despite expectations of greater weekend irregularity. We revisit this phenomenon in detail in Section~\ref{sec:ablation}.

\vspace{4pt}
\noindent\textbf{Distribution Analysis.} Figure~\ref{fig4} presents kernel density estimates (KDE) of predicted versus ground-truth departure times. Historical baselines exhibit high concentration around global averages, particularly in morning and afternoon periods, failing to capture contextual routines relevant to departure. In contrast, our model’s distribution closely tracks the ground-truth curve, demonstrating temporal adaptability despite minor deviations. A slight bias toward earlier predictions is observed, primarily due to the decision threshold $p=0.1$. This suggests that adjusting the $p$ improves practicality of \ours by providing a buffer time before departure.

\begin{table}[t]
    \centering
    \resizebox{\linewidth}{!}{%
    \begin{tabular}{c|l|c|c|c}
        \textbf{Data} & \textbf{Models} & \textbf{All days} & \textbf{Weekdays} & \textbf{Weekends} \\
        \hline
         & MLR & 2.90 & 2.93 & 2.82 \\
         & SVR & 2.57 & 2.52 & 2.63 \\
        History & LGBM & 2.58 & 2.54 & 2.59 \\
         & FTT & 2.74 & 2.72 & 2.76 \\
         & iTransformer & 2.61 & 2.59 & 2.65 \\
        \hline
         & LR & 2.67 & 2.61 & 2.79 \\
         & SVC & 2.66 & 2.64 & 2.70 \\
        Passive & LGBM & 2.72 & 2.81 & 2.65 \\
         & FTT & 2.70 & 2.63 & 2.84 \\
         & iTransformer & 2.59 & 2.56 & 2.65 \\
        \hline
        Passive & \textbf{Ours} & \textbf{2.20} & \textbf{2.26} & \textbf{2.07} \\
    \end{tabular}
    }
    \caption{Performance results (MAE) for baselines and the proposed model across day categories.}
    \label{tab:baseline_comparison}
\end{table}

\begin{table}[t]
    \centering
    \resizebox{\linewidth}{!}{%
    \begin{tabular}{l|c|c|c}
        \textbf{Variants} & \textbf{All days} & \textbf{Weekdays} & \textbf{Weekends} \\
        \hline
        w/o context & 4.47 ± 0.18 & 4.88 ± 0.18 & 3.55 ± 0.16 \\
        w/o PE & 4.25 ± 1.74 & 4.19 ± 1.54 & 4.40 ± 2.23 \\
        w/o time & 3.01 ± 0.67 & 2.96 ± 0.60 & 3.10 ± 0.85 \\
        w/o DoW & 2.64 ± 0.60 & 2.63 ± 0.45 & 2.66 ± 0.94 \\
        w/o alpha-fusion & 2.55 ± 0.42 & 2.58 ± 0.32 & 2.49 ± 0.66 \\
        w/o t-scale \& GSS & 2.36 ± 0.04 & 2.42 ± 0.03 & 2.24 ± 0.09 \\
        \hline
        \textbf{Full Model} & \textbf{2.20 ± 0.13} & \textbf{2.26 ± 0.13} & \textbf{2.07 ± 0.17} \\
    \end{tabular}
    }
    \caption{Ablation study on design choices. Variants are created by removing components from the Full Model.}
    \label{tab:ablation_study}
\end{table}

\begin{figure*}[t]
\centering
\includegraphics[width=1\textwidth]{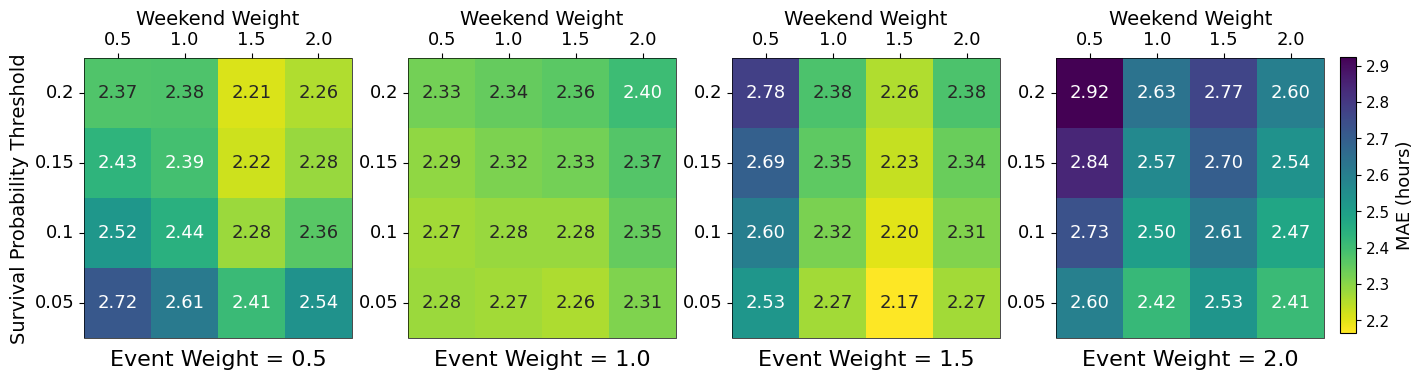} 
\caption{The impact of hyperparameters: event weight $\omega_e$, weekend weight $\omega_w$, and survival probability threshold $p$ on the departure time prediction accuracy.}
\label{fig5}
\end{figure*}

\subsection{Ablation Study}
\label{sec:ablation}
To evaluate the contribution of key design components, we conduct an ablation study along three dimensions: (1) input features, (2) architectural and regularization strategies, and (3) loss design. Table~\ref{tab:ablation_study} presents MAE (hours) for all days, weekdays, and weekends across these variants.

\vspace{4pt}
\noindent\textbf{Feature Contributions.} Removing contextual features (w/o context) leads to the largest performance drop (4.47 hours), underscoring the critical role of behavioral and environmental indicators in capturing pre-departure routines. Removing absolute time features (w/o time) results in moderate degradation (3.01 hours), indicating that temporal position cues are important for capturing regular temporal patterns within each day. Excluding day-of-week features (w/o DoW) slightly worsens performance (2.64 hours), highlighting their usefulness in capturing weekly rhythms.

\vspace{4pt}
\noindent\textbf{Architecture and Regularization.} Removing positional encoding (w/o PE) substantially increases error (4.25 hours), confirming its necessity for temporal order modeling. Disabling alpha-fusion (w/o alpha-fusion) also degrades accuracy (2.55 hours), highlighting the benefit of balancing contextual and temporal representations.

\vspace{4pt}
\noindent\textbf{Loss Design and Synergistic Regularization.} Removing both time-scale and GSS (w/o t-scale \& GSS) increases MAE to 2.36 hours, confirming their synergistic benefit. Time-scale improves temporal calibration across discretized time grids, while GSS provides soft supervision that mitigates over-penalization of near-boundary timing shifts.

Overall, the ablation analysis confirms that all components jointly contribute to performance, with contextual features and positional encoding emerging as the most influential factors. Notably, several configurations exhibit lower MAE on weekends than on weekdays, contrary to the expectation of greater weekend irregularity. A plausible explanation lies in the dataset’s collection period (2021--2022), which coincided with COVID-19 and post-pandemic transitions, where hybrid work schedules and flexible weekday routines introduced more variability than weekends.

\subsection{Hyperparameter Sensitivity Analysis}
\label{sec:hyperparam}
We examine the sensitivity of our model to three key hyperparameters: event weight $\omega_e$, weekend weight $\omega_w$, and the survival probability threshold $p$. Figure~\ref{fig5} presents the MAE heatmaps across different combinations.

\vspace{4pt} 
\noindent\textbf{Effect of Event Weight.} Each subplot corresponds to a fixed value of $\omega_e$. Medium values (1.0–1.5) yield the best MAE overall, especially when paired with moderate $\omega_w$. Small values weaken the event signal, while large values overemphasize the event point and harm calibration. Notably, the model's stable performance at $\omega_e = 1$ suggests that balancing survival and failure signals promotes robust training and stable outcomes. The best results are achieved when $\omega_e = 1.5$ is combined with $\omega_w = 1.5$ and a small $p$.

\vspace{4pt}
\noindent\textbf{Effect of Weekend Weight.} Small $\omega_w$ weakens weekend learning, causing the model to fall back on regular weekday patterns and making weekend cues less reliable. With large $\omega_e$ and $p$, these weak early cues combined with a softened threshold lead to unstable early predictions. With small $\omega_e$ and $p$, the event signal is underemphasized and the strict threshold delays detection, producing late predictions. Thus, small $\omega_w$ consistently harms weekend performance, while $\omega_e$ and $p$ determine whether errors occur early or late.

\vspace{4pt}
\noindent\textbf{Effect of Survival Probability Threshold.} $p$ shows a consistent MAE pattern centered around $\omega_e = 1$, where survival and failure signals are balanced. When $\omega_e = 1.0\text{–}2.0$, smaller $p$ improves accuracy since the model learns sharper survival drops near the event. When $\omega_e = 0.5$, smaller $p$ delays detection and worsens performance. In short, strict $p$ works well when the event signal is emphasized.

\vspace{4pt}
\noindent\textbf{Interactions and Practical Settings.} Overall, moderate $\omega_e$ and $\omega_w$ (around 1.5) combined with a small $p$ yield the most reliable performance. Small $\omega_w$ consistently harms weekend predictions, while extreme $\omega_e$ or $p$ shifts the model toward unstable early or late detections. Using moderate weights and a strict threshold avoids these issues and provides accurate, stable real-time departure time predictions.

\subsection{Personalization}
\label{sec:personalization}
We further evaluate user-specific adaptation through fine-tuning. All global model parameters remain frozen, and only the last Transformer layer and output layer are updated using user-specific data. Personalization yields a modest overall improvement (MAE: 2.20 → 2.13 hours), with larger gains on weekends (2.07 → 1.85 hours) than weekdays (2.26 → 2.23 hours). This trend aligns with earlier observations of lower weekend MAE, likely influenced by COVID-19 schedule shifts and the effect of $\omega_w$ on weekend variability. Additional analysis of intra-user variability, its impact, and a cold-start strategy is provided in the supplementary material. Future improvements may incorporate meta-learning or adaptive regularization to better balance global patterns with individual behaviors, and richer contextual signals from large language models may further enhance personalization. 

\section{Conclusion}
\label{sec:conclusion}
This paper presents a Transformer-based TTD modeling framework as the foundation for \ours, a strategy aimed at reducing battery degradation and extending EV battery lifetime. Unlike prior work that relies on historical patterns, our approach uses real-time contextual signals from unobtrusive smartphone sensing and models departure as a survival process on discretized time grids. In a large-scale in-the-wild study with 93 participants and 3,906 daily sequences, our method outperforms historical and context-aware baselines, reducing MAE by 14.4\%. Ablation studies highlight the role of design considerations in robust temporal modeling, while fine-tuning enables further personalization. The proposed dataset and framework establish a benchmark for real-time departure prediction in sustainable EV smart charging, supporting both battery longevity and carbon neutrality.

\section{Supplementary Material}

\subsection{Field Study and Dataset}
\label{sec:dataset}

We conducted a large-scale field study under our Institutional Review Board (IRB) approval, following strict privacy and security guidelines. This section summarizes participant recruitment, data collection, and ground-truth extraction~\cite{lee2024extending} for additional details and illustrations. Due to space and capacity constraints, we provide a small sample of the preprocessed dataset along with evaluation code in the supplementary material. The full dataset will be released upon acceptance of the paper).

\subsubsection{User Study}
The study aimed to explore the feasibility of contextual features for predicting vehicle (EV) departure time. Due to the low proportion of EV users, we broadened recruitment to individuals aged 18--69 (male: 234, female: 268; mean age: 28.5, SD: 10.6), assuming that departure patterns between EVs and general vehicles (internal combustion engine vehicles) do not differ significantly. Participants were recruited via social media, universities, online communities, and public advertisements in the Seoul metropolitan area. Recruitment occurred in three phases over 13 months (May 31, 2021--June 27, 2022), with observation concluding on July 5, 2022.

Participants installed our custom Android application, EV Analyzer (EVA), for passive sensing data collection in-the-wild. Informed consent was obtained, including user study details and privacy protections. Participants received up to \$50 for participation. All data were encrypted (AES-256-CBC) and stored on a secure server; identifiers were anonymized. 

\subsubsection{Data Collection}
We developed an Android mobile application, EVA (EV Analyzer), for continuous and event-based passive data collection. EVA collects nine passive sensing modalities representing behavioral and environmental contexts: activity transition, steps, significant motion, calls, screen state, unlock state, app usage, ambient light, and ambient sound. Continuous sensing-based data (\eg light, sound) were sampled every 5 minutes to balance resolution and battery consumption, while event-based data (\eg activity, app usage) were recorded upon state changes. 

\subsubsection{Ground Truth}
Ground-truth departure times were derived from the activity transition sensor. Specifically, the first occurrence of the activity type \texttt{IN\_VEHICLE} after 4:00 AM within each daily sequence was defined as the vehicle departure time. This approach captures the earliest confirmed vehicle-use event for the day, aligning with typical morning departure behavior. 

\subsection{Data Preprocessing}
We implemented a structured data preprocessing pipeline encompassing participant filtering, sensor data cleaning, feature engineering, and temporal encoding to ensure reliability of our work.

\subsubsection{Data Cleaning}
Mobile passive sensing data collected in-the-wild often contain missing values, redundant records, or unrealistic measurements due to device and transmission constraints. We applied the following steps:

\begin{itemize}
    \item \textbf{Participant filtering:} Participants were retained only if they satisfied \emph{all} of the following conditions: (1) at least one data entry was available across all sensors, (2) data were collected for a minimum of two months, (3) at least 42 valid daily samples existed within this two-month period, and (4) no more than one sample was missing for both weekdays and weekends in any given week. Participants failing any of these conditions were excluded.
    \item \textbf{Temporal integrity check:} Removed duplicate or temporally inconsistent records (\eg reversed timestamps) to preserve chronological order.
    \item \textbf{Outlier removal:} Filtered unrealistic values based on domain thresholds (\eg ambient light 0–10{,}000 lux; sound energy $<0$ dBFS).
\end{itemize}

As a result, the final dataset included 93 participants with at least 42 days each, totaling 3{,}906 daily samples and 94 engineered features across nine modalities.

\subsubsection{Feature Engineering}
\label{sec:section5.2}
Features were extracted from nine passive sensing modalities, capturing behavioral and environmental contexts. Event-based sensors (\eg activity type, steps, screen state) contributed duration, count, ratio, entropy, and elapsed time since first occurrence. Continuous sensors (\eg light, sound) provided statistical descriptors (mean, min, max, std., skewness, kurtosis) within fixed time windows. In total, 94 features were generated (See Table ~\ref{tab:departure_features}). 

\subsubsection{Sliding Window and Temporal Encoding}
\label{sec:section5.3}
To capture temporal dynamics, we applied a sliding window approach, extracting features over the optimal window per individual and updating every 5 minutes. Additional temporal features included current hour, binary day-of-week (weekday/weekend, holiday) indicators, providing time contexts for the Transformer-based TTD model.

\begin{table*}[ht!]
\centering
\small
\caption{Extracted features from passive sensing data, comprising 94 features in total.}
\label{tab:departure_features}
\setlength\tabcolsep{6pt}
\begin{tabular}{p{4cm} p{6.2cm} p{5.8cm}}
\toprule
\textbf{Data Type} & \textbf{Extracted Features} & \textbf{Relevance to Departure Detection} \\
\midrule
\multicolumn{3}{l}{\textbf{Behavioral Data}} \\
\midrule
Activity Type (still, walking, running, on\_bicycle) &
Durations and counts of 4 activity types; ratios of each type to total; elapsed time since first occurrence & \\
\cmidrule{1-2}
Step &
Count of steps; step duration std.; elapsed time since first step & 
\multirow{3}{=}{Physical activities (movements) as indicators of departure} \\
\cmidrule{1-2}
Significant Motion &
Count of significant motion; elapsed time since last and first motion & \\
\midrule
Call (in, out, missed) &
Durations and counts of 3 call types; ratios (in/out, per total); elapsed time since first occurrence &
Social (call) activities as indicators of departure \\
\midrule
Screen State &
Durations of screen-on; screen state counts and short-on events; elapsed time since first screen-on; entropy and normalized entropy & \\
\cmidrule{1-2}
Unlock State &
Durations of unlock states; unlock/lock counts; elapsed time since first unlock; entropy and normalized entropy & 
\multirow{3}{=}{Smartphone usage behaviors as indicators of \\ departure}\\
\cmidrule{1-2}
App Usage (\footnotesize{social, communication, lifestyle, entertainment, health, book, video, finance, photography, productivity}) &
Durations and counts of 10 app types; elapsed time since first usage; entropy and normalized entropy & \\
\midrule
\multicolumn{3}{l}{\textbf{Environmental Data}} \\
\midrule
Ambient Light &
Count of complete darkness; min, max, mean, std., kurtosis, skewness of brightness (lux) &
Light intensity and variations as indicators of departure \\
\midrule
Ambient Sound &
Min, max, mean, std., kurtosis, skewness of sound energy (dBFS) &
Sound intensity and variations as indicators of departure \\
\bottomrule
\end{tabular}
\end{table*}

\subsection{Correlation Analysis}
\begin{figure*}[t]
\centering
\includegraphics[width=0.8\textwidth]{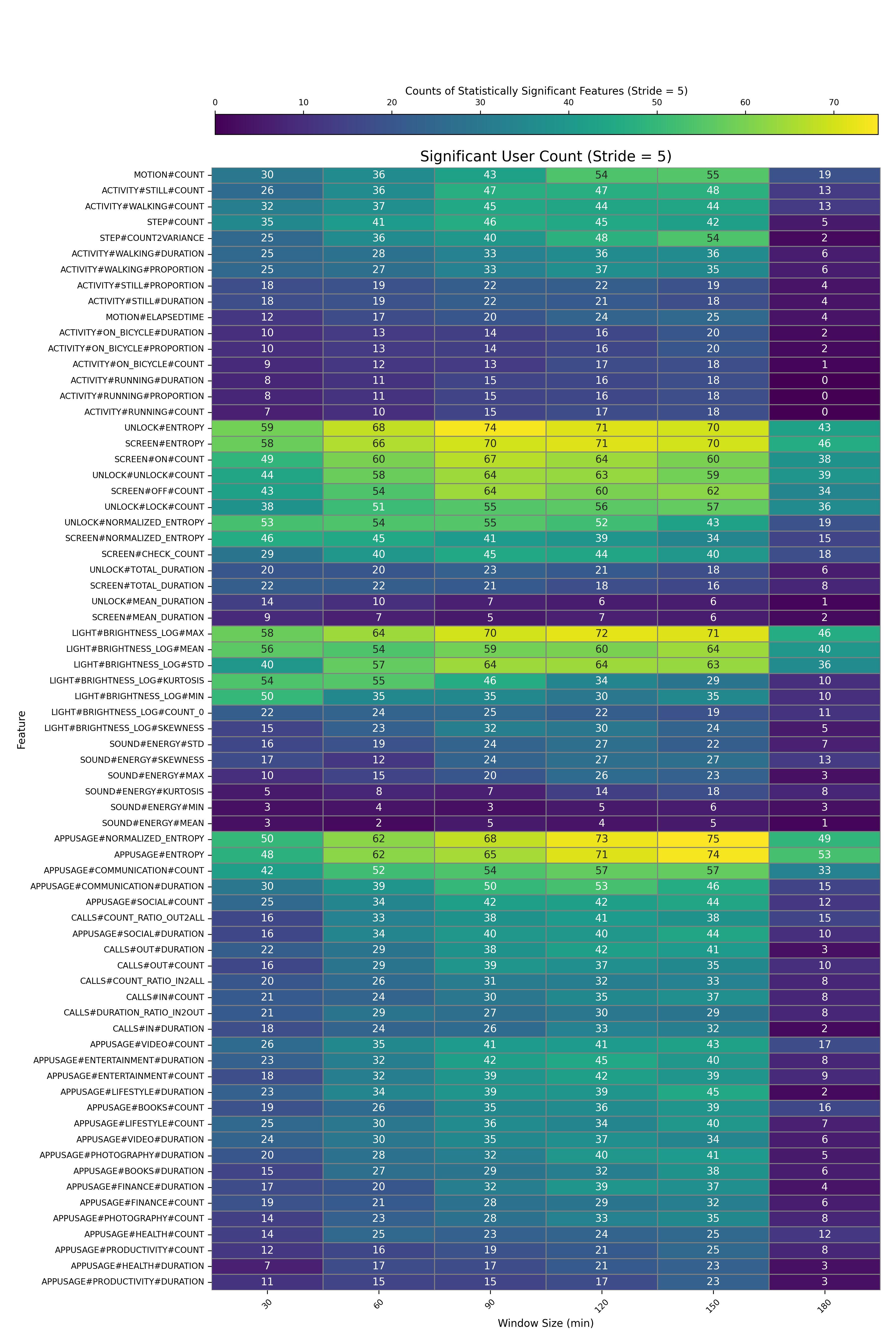} 
\caption{Heatmap showing statistical significance ($p<0.05$) of 73 contextual features across time windows (1–3 hours, 30-min stride). Features such as unlock state, screen state, app usage, and light exhibit highest significance, particularly at 2 hours, which was selected as the optimal window for feature extraction.}
\label{apdx_fig1}
\end{figure*}

We first conducted a correlation analysis on 73 contextual passive features, which were extracted at each time step $t$ as contextual embeddings corresponding to tokens in the daily input sequence. To distinguish patterns between departure events and non-departure ranges prior to the events, we used the point-biserial correlation coefficient.

Figure~\ref{apdx_fig1} shows a heatmap comparing multiple time windows (1–3 hours, 30-minute stride, total of 6 cases) sorted by contextual categories. The heatmap indicates the number of participants for whom each feature achieved statistical significance ($p < 0.05$) across windows. Notably, features such as unlock state, screen state, general app usage, communication app usage, and light exhibited strong significance, particularly at the 2-hour window. Based on these results, we selected a 2-hour window as the optimal context length for pre-departure detection.

In addition to contextual features, we use the first occurrence time of each of seven event-based sensor modalities (and subcategories) as temporal features. These modalities include activity type (running, walking, on\_bicycle, still), motion, step, screen state, unlock state, app usage (social, communication, lifestyle, entertainment, health, book, video, finance, photography, productivity), and calls (in, out, missed),
yielding 21 hourly features in total. Due to feature distribution characteristics, we applied Spearman rank correlation. 

\begin{figure*}[t]
\centering
\includegraphics[width=0.9\textwidth]{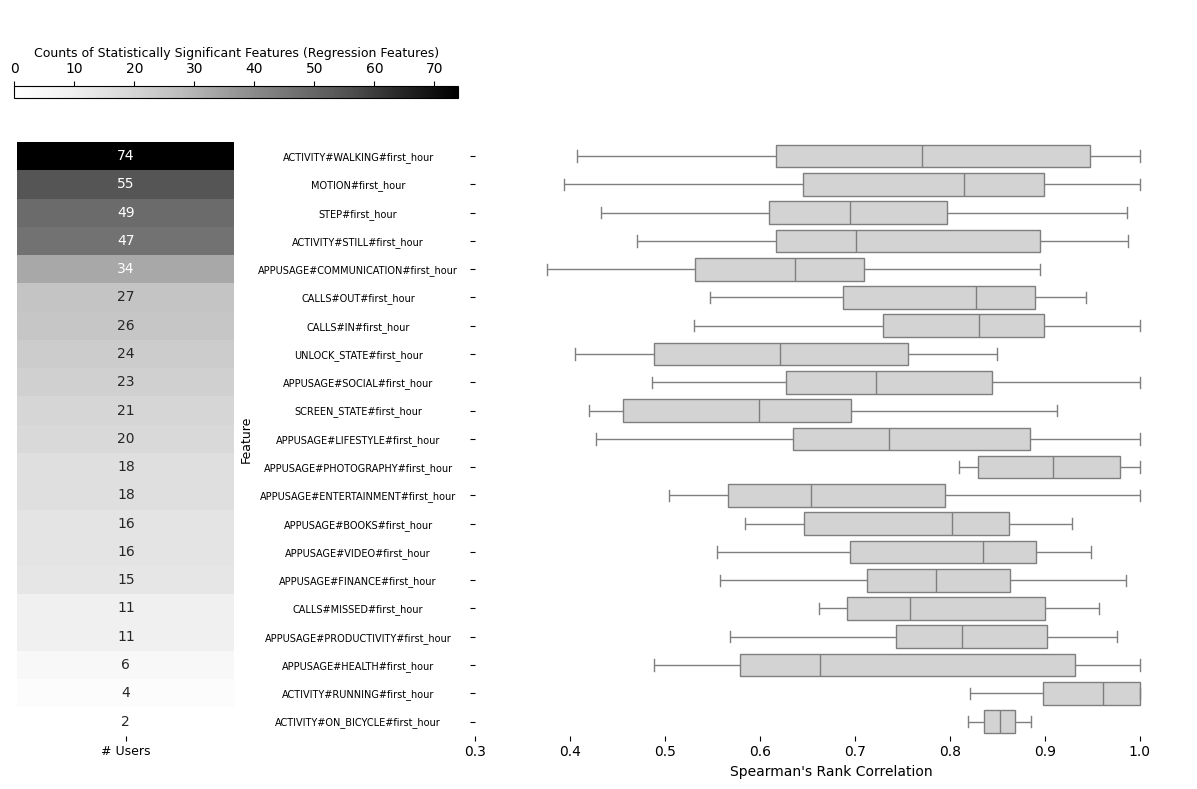} 
\caption{Statistical significance ($p<0.05$) and correlation distribution of 21 event-based temporal features. General physical activity features (\eg walking, motion, step) show broad significance, while intensive activities (\eg running, cycling) exhibit localized effects, highlighting inter-individual variability and the need for personalized modeling.}
\label{apdx_fig2}
\end{figure*}

Figure~\ref{apdx_fig2} presents (left) a heatmap of statistical significance ($p < 0.05$) and (right) boxplots of correlation coefficients, with features ranked by the number of participants exhibiting significance. The heatmap shows participant counts per feature achieving significance, while boxplots depict coefficient ranges, reflecting inter-individual heterogeneity. Notably, general physical activity features (\eg walking, motion, step) consistently demonstrate significance across users, whereas intensive activities (\eg running, cycling) exhibit localized significance. These findings indicate that while common behaviors reflect general pre-departure routines, personalized modeling remains essential.

\subsection{Implementation Details}
\noindent
We tuned model and training hyperparameters using a grid search with 5-fold cross-validation on the training set to mitigate overfitting. The hyperparameter search space was adapted from the Transformer-based deep survival analysis framework~\cite{hu2021transformer}, but we reduced the size of the search ranges and focused on small values to reflect the limited size of our dataset (3,906 samples) and ensure model compactness. The final search space was:

\begin{itemize}
    \item Number of Transformer layers ($\text{num\_layers}$): \{3, 4\}
    \item Number of attention heads ($\text{n\_head}$): \{1, 2\}
    \item Hidden dimension ($d_{\text{model}}$): \{32, 64\} 
    \item Dropout rate ($\text{dropout}$): \{0.1\} (fixed)
    \item Temporal dropout ($\text{dropout\_time}$): \{0.1, 0.2\}
    \item Learning rate (LR): \{1e-4, 1e-3\}
    \item Weight decay (L2 regularization): \{5e-4, 1e-3\}
\end{itemize}

\noindent
The optimal configuration was selected based on the average validation performance across folds. The final generalized model employs three Transformer layers, $d_{\text{model}} = 32$, one attention head, $\text{dropout}=0.1$, and is trained with the Adam optimizer (Kingma and Ba, 2015) using a learning rate of $10^{-3}$ and weight decay of $5 \times 10^{-4}$. Hyperparameter tuning was conducted only for the generalized (global) model; the personalized model uses fine-tuning strategies, as summarized in Table~\ref{table:global_vs_personalized}, which provides a detailed comparison of architecture and training configurations for both models.

\begin{table}[ht!]
\centering
\caption{Hyperparameter settings for generalized and personalized models. Hyperparameter tuning was applied only for the generalized model.}
\label{table:global_vs_personalized}
\setlength\tabcolsep{4.5pt}
\resizebox{0.48\textwidth}{!}{
\begin{tabular}{llcc}
\toprule
\textbf{Category} & \textbf{Parameter} & \textbf{Generalized Model} & \textbf{Personalized Model} \\
\midrule
\multicolumn{4}{l}{\textbf{Architecture (Fixed)}} \\
\midrule
& Input dimension & \multicolumn{2}{c}{94} \\
& Sequence length & \multicolumn{2}{c}{265} \\
& Transformer layers & \multicolumn{2}{c}{3} \\
& Embedding dimension ($d_{model}$) & \multicolumn{2}{c}{32} \\
& Attention heads & \multicolumn{2}{c}{1} \\
& Positional encoding & \multicolumn{2}{c}{Enabled} \\
\midrule
\multicolumn{4}{l}{\textbf{Training (Shared)}} \\
\midrule
& Weekend weight & \multicolumn{2}{c}{1.5} \\
& Event weight & \multicolumn{2}{c}{1.5} \\
& Survival probability threshold & \multicolumn{2}{c}{0.1} \\
\midrule
\multicolumn{4}{l}{\textbf{Training (Distinct)}} \\
\midrule
& Dropout & 0.1 & 0.0 \\
& Temporal dropout & 0.2 & 0.2 \\
& Learning rate (LR) & 1e-3 & 5e-4 \\
& Batch size & 64 & 4 \\
& Epochs & 100 & 50 \\
& Early stopping patience & 5 & 3 \\
& Weight decay / L1 & 5e-4 / 0.0 & 0.0 / 1e-3 \\
\midrule
\multicolumn{4}{l}{\textbf{Fine-tuning}} \\
\midrule
& $k$ (number of last layers) & -- & 1 \\
\midrule
\multicolumn{4}{l}{\textbf{Hardware}} \\
\midrule
& GPU & RTX 4090 & -- \\
& Time & 7h & -- \\
\bottomrule
\end{tabular}
}
\end{table}

\subsection{Survival Probability Analysis}
During inference, survival probability estimates are updated in real time as new contextual observations are streamed. At each interval $t$, the input token comprising contextual and temporal features is processed, and the predicted hazard complement $\hat{q}(t|\mathbf{X})$ updates the cumulative survival probability $S(t|\mathbf{X})$. Detection occurs when $S(t|\mathbf{X})$ falls below a predefined threshold (\eg 0.1).

\begin{figure*}[t]
\centering

\begin{minipage}[t]{0.32\textwidth}
    \centering
    \includegraphics[width=\linewidth]{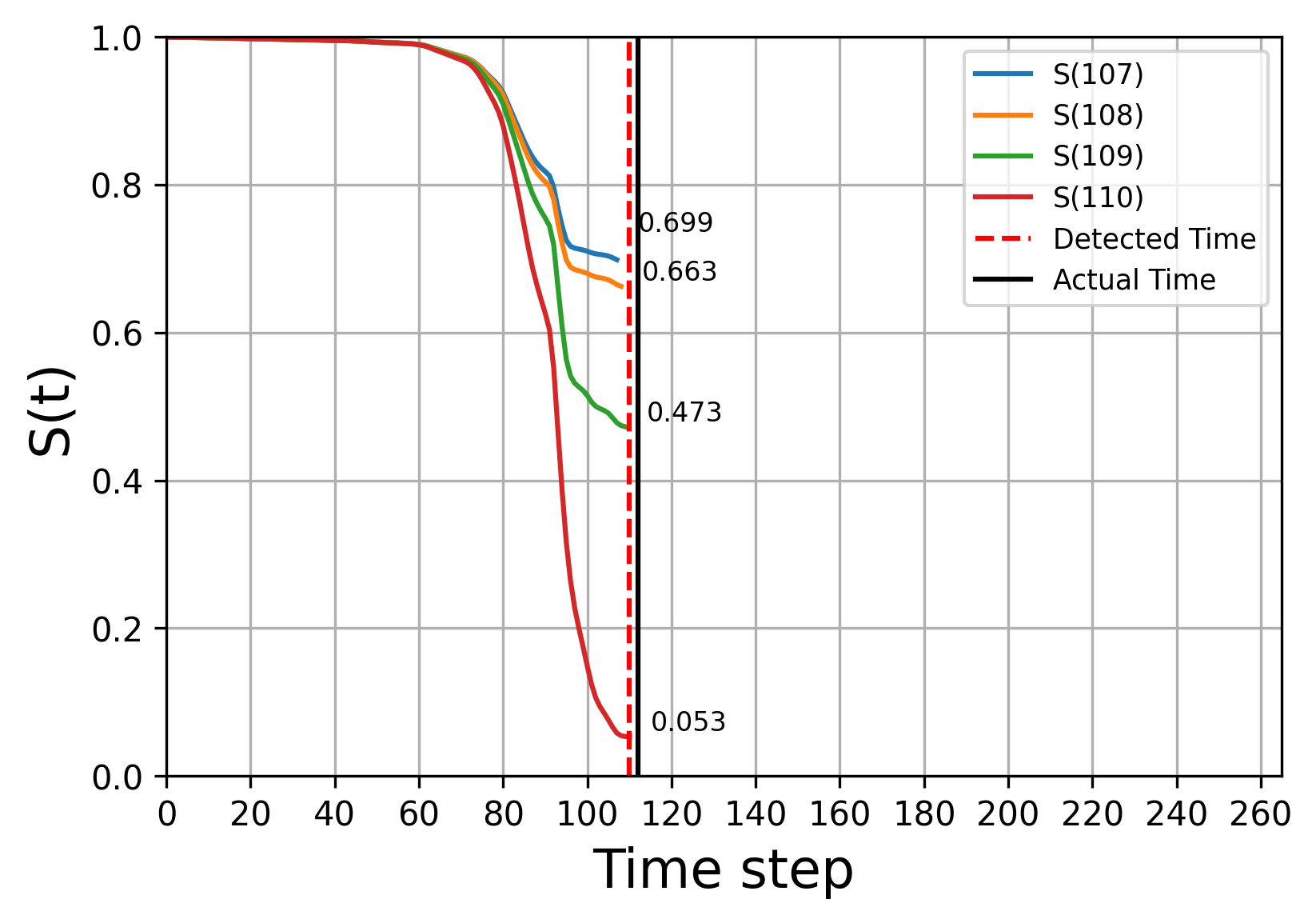}
    \caption*{\textbf{(a)} Sharp decrease}
\end{minipage}
\hfill
\begin{minipage}[t]{0.32\textwidth}
    \centering
    \includegraphics[width=\linewidth]{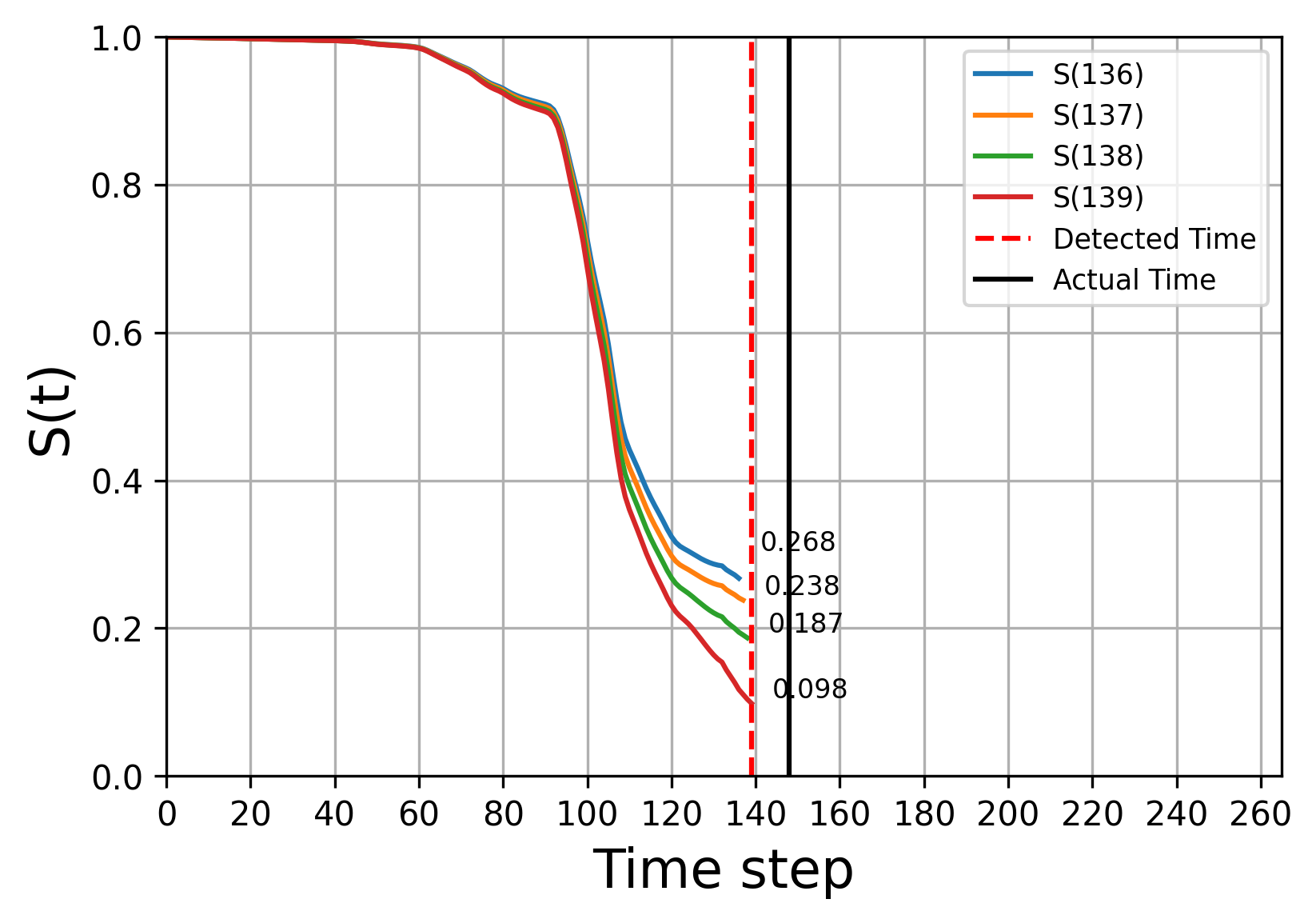}
    \caption*{\textbf{(b)} Gradual decline}
\end{minipage}
\hfill
\begin{minipage}[t]{0.32\textwidth}
    \centering
    \includegraphics[width=\linewidth]{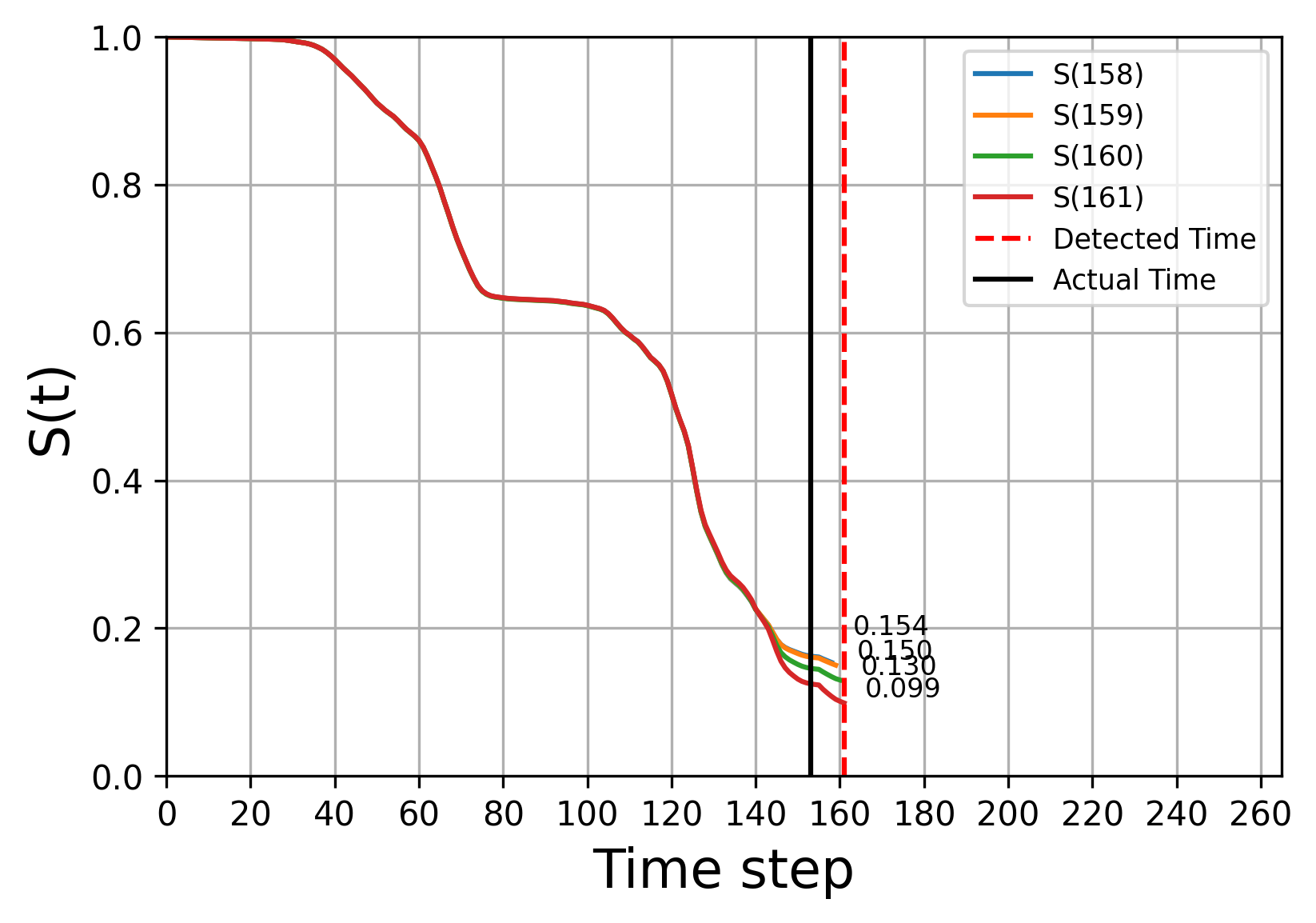}
    \caption*{\textbf{(c)} Step-wise}
\end{minipage}

\vspace{1em}
\caption{
Evolution of survival probability $S(t)$ during inference for three representative user patterns.
\textbf{(a)} Sharp decrease triggered by sudden contextual cues.
\textbf{(b)} Gradual decline driven by both temporal and contextual indicators.
\textbf{(c)} Step-wise pattern with multiple significant drops preceding detection.
These patterns demonstrate the model’s adaptability in real-time inference and its ability to capture heterogeneous behavioral dynamics.
}
\label{fig:supp_survival}
\end{figure*}

To illustrate this dynamic process, Figure~\ref{fig:supp_survival} visualizes survival probability evolution for a representative user across three typical patterns. Each subplot shows the survival probability curve at the detected step~$t$, the preceding three curves ($t\!-\!1$ to $t\!-\!3$), and vertical lines indicating the actual and detected departure times, highlighting the real-time update mechanism.

\begin{itemize}
    \item \textbf{Sharp Decrease (Figure 8a):} The survival probability curve remains high and then drops sharply once critical contextual cues emerge. This indicates the model identifies strong departure signals at the moment, triggering detection.
    \item \textbf{Gradual Decline (Figure 8b):} Survival probability decreases smoothly over several intervals, suggesting temporal features and contextual patterns jointly signal an impending departure.
    \item \textbf{Step-wise Pattern (Figure 8c):} The curve exhibits multiple sharp declines before the event, reflecting intermittent strong cues (\eg preparatory activities before departure) that increase departure likelihood incrementally.
\end{itemize}

For a deeper understanding of the factors driving changes in $S(t|\mathbf{X})$ leading to departure detection, we performed an attribution analysis using Integrated Gradients (IG)~\cite{sundararajan2017axiomatic}, a widely adopted gradient-based feature attribution method. Specifically, we computed IG scores for the top 10 input features comprising both contextual and temporal components between the time step when detection was triggered ($t$) and the preceding step ($t-1$). The analysis revealed that general physical activity indicators (\eg walking, step counts) and localized behaviors such as app usage (\eg reading, entertainment) were among the most influential factors in reducing survival probability. These findings are consistent with the classification analysis results, highlighting that both generalized activity patterns and personalized routines serve as significant pre-departure cues.

\subsection{Impact of Individual Departure Time Variability}
Departure time patterns differ substantially across individuals, influencing prediction difficulty. Users with highly variable departure schedules exhibit greater uncertainty, making accurate detection more challenging. To quantify this effect, we computed the standard deviation (STD) of daily departure times for each participant and compared it to the corresponding MAE of the prediction model. 

\begin{figure}[t]
\centering
\includegraphics[width=0.4\textwidth]{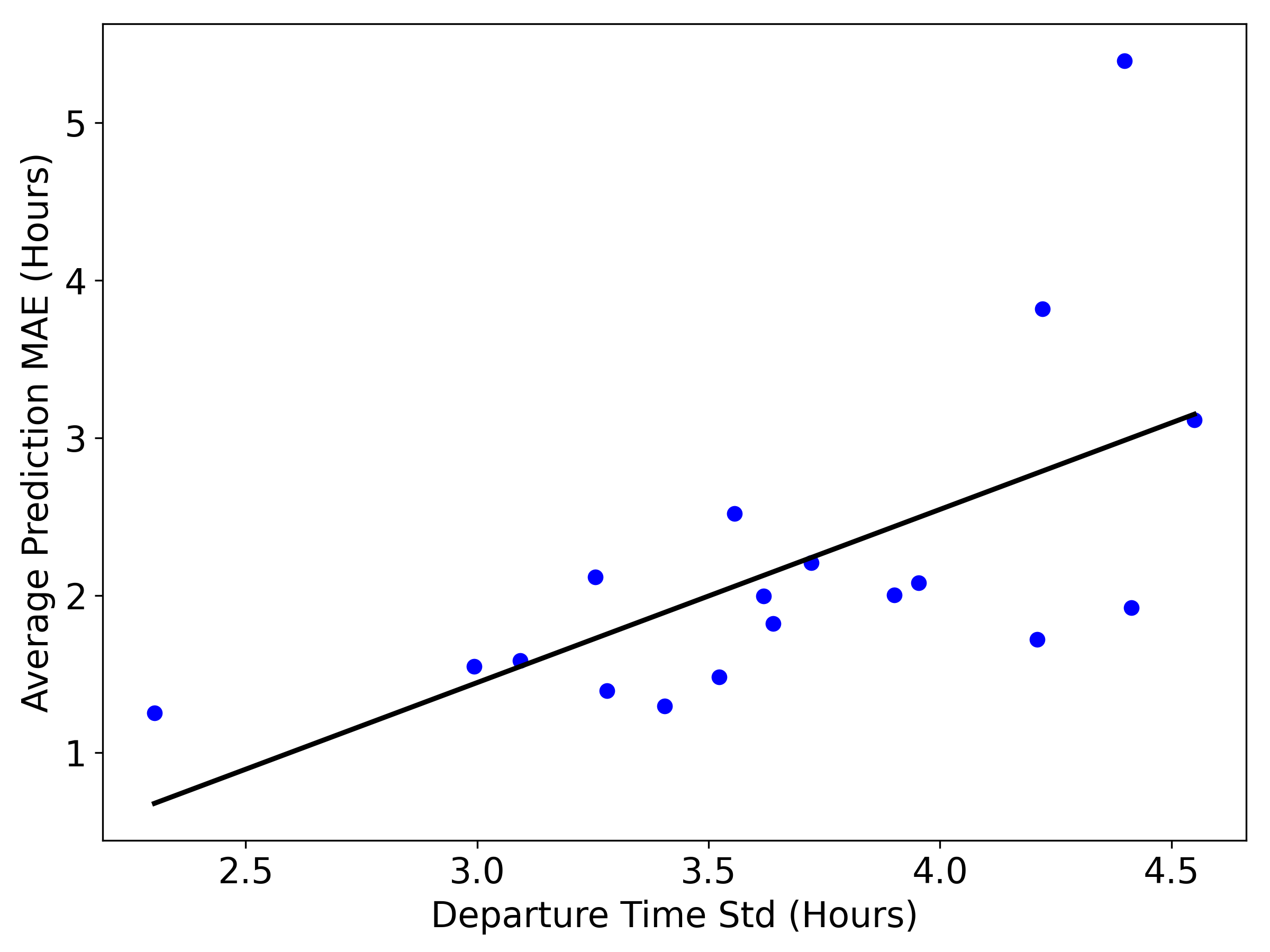} 
\caption{Relationship between individual departure time variability and prediction error. Each point represents a participant, where the $x$-axis shows the standard deviation of departure time and the $y$-axis shows the corresponding MAE. A statistically significant positive correlation was found ($r = 0.62$, $p = 0.0063$), suggesting that users with higher variability in departure time tend to have larger prediction errors.}
\label{apdx_fig_4}
\end{figure}

Figure ~\ref{apdx_fig_4} shows a clear positive correlation with statistical significance ($r = 0.62$, $p = 0.0063$); individuals with larger STD values tend to have higher MAE, indicating that irregular departure routines reduce prediction accuracy. This variability highlights the importance of advanced personalized modeling strategies to address user-specific behavioral heterogeneity.

\subsection{User Acceptance and Practicality}
\subsubsection{User Acceptance}
As \ours{} implements real-time departure detection, it relies on the quality of continuously streamed passive features from smartphones. This dependency raises critical considerations for user acceptance, particularly regarding privacy and sustained engagement. To examine these factors, we conducted an preliminary, longitudinal study on passive sensing data collection using the UTAUT2 framework under IRB approval ~\cite{kondoro2025investigating}. Results indicate that privacy concerns significantly predict long-term behavioral intention and moderate constructs such as trust and effort expectancy. While habit strength declined, behavioral intention increased, suggesting a shift toward deliberate engagement. To further support practical deployment and social impact, we are conducting an extended study based on Technology Acceptance Model (TAM)~\cite{venkatesh2000theoretical}, focusing on privacy, perception-aware design, and ethical adoption.

\subsubsection{Inference Time and Throughput}
We assessed the real-time feasibility of \ours{} by measuring inference latency and throughput on an RTX 4090 GPU. Table~\ref{tab:inference} reports results for generalized and personalized models. The generalized model achieved $2.13 \pm 0.03$ ms per inference (469.16 samples/s), while personalized models averaged $2.68 \pm 0.05$ ms (373.70 samples/s), about 25.8\% slower. Despite this overhead, both configurations operate within millisecond-level latency, confirming the practicality of real-time deployment.

\begin{table}[t]
\centering
\scriptsize
\caption{Inference time and throughput for generalized and personalized models. Results are reported as mean $\pm$ standard deviation across runs. Personalized models show increased latency compared to the generalized model but remain suitable for real-time operation.}
\label{tab:inference}
\begin{tabular}{lcc}
\toprule
\textbf{Model Type} & \textbf{Inference Time (ms)} & \textbf{Throughput (samples/s)} \\
\midrule
Generalized Model & $2.13 \pm 0.03$ & 469.16 \\
Personalized Models (avg) & $2.68 \pm 0.05$ & 373.70 \\
\bottomrule
\end{tabular}
\end{table}

\subsubsection{Battery Consumption} The practicality of \ours{} depends on balancing continuous sensing with resource constraints on user devices (smartphones). Real-time operation begins when the SOC reaches 80\% and continues until departure is detected, requiring streaming data collection and processing of multiple smartphone sensors. This can increase battery drain and pose adoption challenges. To mitigate this, the EVA app minimizes impact by running in the background and suspending sensing when the phone battery drops below 20\%. For large-scale deployment, computationally intensive tasks such as model inference will be offloaded to a cloud-based BMS, enabling real-time decision-making while reducing on-device load.

\subsection{Related Work on Deep Survival Analysis}

Recent advances in deep survival analysis have introduced recurrent, attention-based, and Transformer-driven architectures, along with Bayesian and generative formulations, to better capture complex hazard dynamics. These models often leverage full-sequence covariate trajectories, jointly model covariate–event interactions, or incorporate auxiliary objectives such as counterfactual prediction, stability, or fairness to enhance performance in conventional survival forecasting tasks.

However, as summarized in Table~\ref{tab:baseline_summary}, the underlying design assumptions of these methods are misaligned with the requirements of our problem setting. Most models (i) treat covariates as endogenous or fixed, which conflicts with our use of exogenous, passively observed, and time-varying behavioral features; (ii) assume access to complete trajectories at inference time, violating our real-time detection; or (iii) depend on predefined priors, kernel structures, or generative decoders that introduce unncessary complexity for a dedicated, single-event detection task. 

In contrast, our task focuses on real-time detection of an individual’s departure event using exogenous, time-varying contextual sequences. Accordingly, we adopt strong sequential baselines (\ie FTTransformer and iTransformer) that directly model temporal and contextual dependencies without relying on hazard-specific assumptions or covariate–event feedback mechanisms. Based on these considerations, we exclude deep survival models such as DeepHit, Dynamic-DeepHit, SurvTRACE, SurvPP, DySurv, TransformerLSR, and fairness- or stability-oriented methods.

\begin{table*}[t]
    \centering
    \renewcommand{\arraystretch}{1.4}
    \setlength{\tabcolsep}{6pt}
    \begin{tabular}{|>{\centering\arraybackslash}p{3cm}|
                        >{\centering\arraybackslash}p{7cm}|
                        >{\centering\arraybackslash}p{7cm}|}
        \hline
        \textbf{Related Work} & \textbf{Method} & \textbf{Rationale for Exclusion} \\
        \hline\hline

        \textbf{DeepHit} \cite{lee2018deephit} &
        Learns the joint distribution of event time and competing risks from static covariates using a deep neural network. &
        Assumes fixed covariates; cannot handle time-varying or real-time contextual dynamics. \\
        \hline

        \textbf{Dynamic-DeepHit} \cite{lee2019dynamic} &
        Extends DeepHit with recurrent layers to jointly model covariate and event processes for dynamic survival prediction. &
        Treats covariates as endogenous variables (joint modeling); unsuitable for exogenous behavioral contexts. \\
        \hline

        \textbf{SurvTRACE} \cite{wang2022survtrace} &
        Transformer-based deep survival model capturing dependencies among static features. &
        Operates on fixed baseline features; lacks temporal sequence modeling and real-time inference capability. \\
        \hline

        \textbf{SurvPP} \cite{kim2023survpp} &
        Bayesian permanental process estimating nonlinear hazard functions via Gaussian process priors. &
        Focuses on kernel-based functional estimation, not representation learning; lacks sequential encoding of behavior. \\
        \hline

        \textbf{DySurv} \cite{mesinovic2024dysurv} &
        Dynamic deep survival model using conditional variational inference for joint modeling of longitudinal covariates and hazard rates. &
        Endogenous joint modeling with batch inference; unsuitable for discrete-time or online event prediction. \\
        \hline

        \textbf{TransformerLSR} \cite{zhang2025transformerlsr} &
        Multi-task encoder–decoder Transformer predicting both future covariates and survival outcomes. &
        Generative and overly complex; includes decoder for covariate forecasting, unnecessary for single-event inference. \\
        \hline

        \textbf{Stable Cox}  \cite{fan2024stablecox} &
        Weighted Cox regression addressing distribution shifts by reweighting samples to identify stable variables. &
        Focuses on causal generalization across environments; not applicable to sequential or contextual hazard inference. \\
        \hline

        \textbf{Fairness DRO} \cite{hu2024fairness} &
        Extends distributionally robust optimization (DRO) to survival analysis to ensure subgroup fairness. &
        Aims at fairness and robustness objectives, not real-time or exogenous time-varying prediction. \\
        \hline
    \end{tabular}
    \caption{Comparison of representative survival models, their methodological focus, and rationale for exclusion as baselines in exogenous, time-varying real-time departure prediction.}
    \label{tab:baseline_summary}
\end{table*}

\section{Ethical Statement}  
This study was conducted in accordance with the principles of the Declaration of Helsinki and was approved by the Institutional Review Board (IRB) of our institution. All participants provided informed consent after being fully informed about the study objectives, procedures, and potential risks. Data collection was performed using a custom mobile application under strict privacy guidelines. Personally identifiable information was never collected, and all data were anonymized prior to analysis using secure hashing methods. Data transmission and storage were protected through AES-256 encryption and secured Linux-based servers. The dataset will be shared exclusively in an anonymized and preprocessed form, strictly for research purposes. These measures ensure participant confidentiality and minimize privacy risks throughout the data lifecycle.
\bibliography{aaai2026.bib}
\end{document}